\documentclass{article}

\usepackage[preprint]{neurips_2026}
\usepackage{caption}
\usepackage{graphicx}
\usepackage{adjustbox}
\usepackage{wrapfig}
\usepackage{todonotes}

\usepackage[utf8]{inputenc} 
\usepackage[T1]{fontenc}    
\usepackage{hyperref}       
\usepackage{url}            
\usepackage{booktabs}       
\usepackage{amsfonts}       
\usepackage{amsmath}        
\usepackage{nicefrac}       
\usepackage{microtype}      
\usepackage[normalem]{ulem}   

\usepackage[table]{xcolor}         
\usepackage{multirow}       
\usepackage{graphicx}       
\usepackage{pifont}
\definecolor{promptback}{rgb}{0.95,0.95,0.95}
\definecolor{promptframe}{rgb}{0.2,0.4,0.7}
\definecolor{prompttitlebg}{rgb}{0.2,0.4,0.7}
\definecolor{promptbody}{rgb}{0.2,0.2,0.2}
\usepackage[most]{tcolorbox}
\usepackage{xcolor}
\usepackage{inconsolata}  
\usepackage{makecell}  
\newtcolorbox{promptbox}[1]{%
  enhanced,
  breakable,
  colback=promptback,
  colframe=promptframe,
  coltitle=white,
  fonttitle=\bfseries\sffamily\small,
  title={#1},
  attach boxed title to top left={yshift=-2.5mm, xshift=4mm},
  boxed title style={
    colback=prompttitlebg,
    sharp corners,
    boxrule=0pt,
    top=1.5mm, bottom=1.5mm, left=2.5mm, right=2.5mm,
  },
  arc=1.5mm,
  boxrule=0.6pt,
  left=4mm, right=4mm, top=5mm, bottom=4mm,
  fontupper=\small\color{promptbody},
  parbox=false,
  before skip=8mm,
  after skip=4mm,
  pad at break*=2mm,
}

 \usepackage{makecell}
 \usepackage{comment}

\title{UnfoldArt: Zero-Shot Recovery of Full Articulated 3D Objects from Text or Image}

\author{%
  Mohamed El Amine Boudjoghra\textsuperscript{1} \quad
  Ivan Laptev\textsuperscript{2} \quad
  Angela Dai\textsuperscript{1} \\[0.5em]
  \textsuperscript{1}Technical University of Munich \\
  \textsuperscript{2}Mohamed Bin Zayed University of Artificial Intelligence \\
}

\begin{document}

\maketitle
\vspace{-1.2cm}

\begin{center}
  \includegraphics[
  width=\textwidth,
  trim={0.00\textwidth{} 0.15\textheight{} 0.00\textwidth{} 0.15\textheight{}},
  clip
]{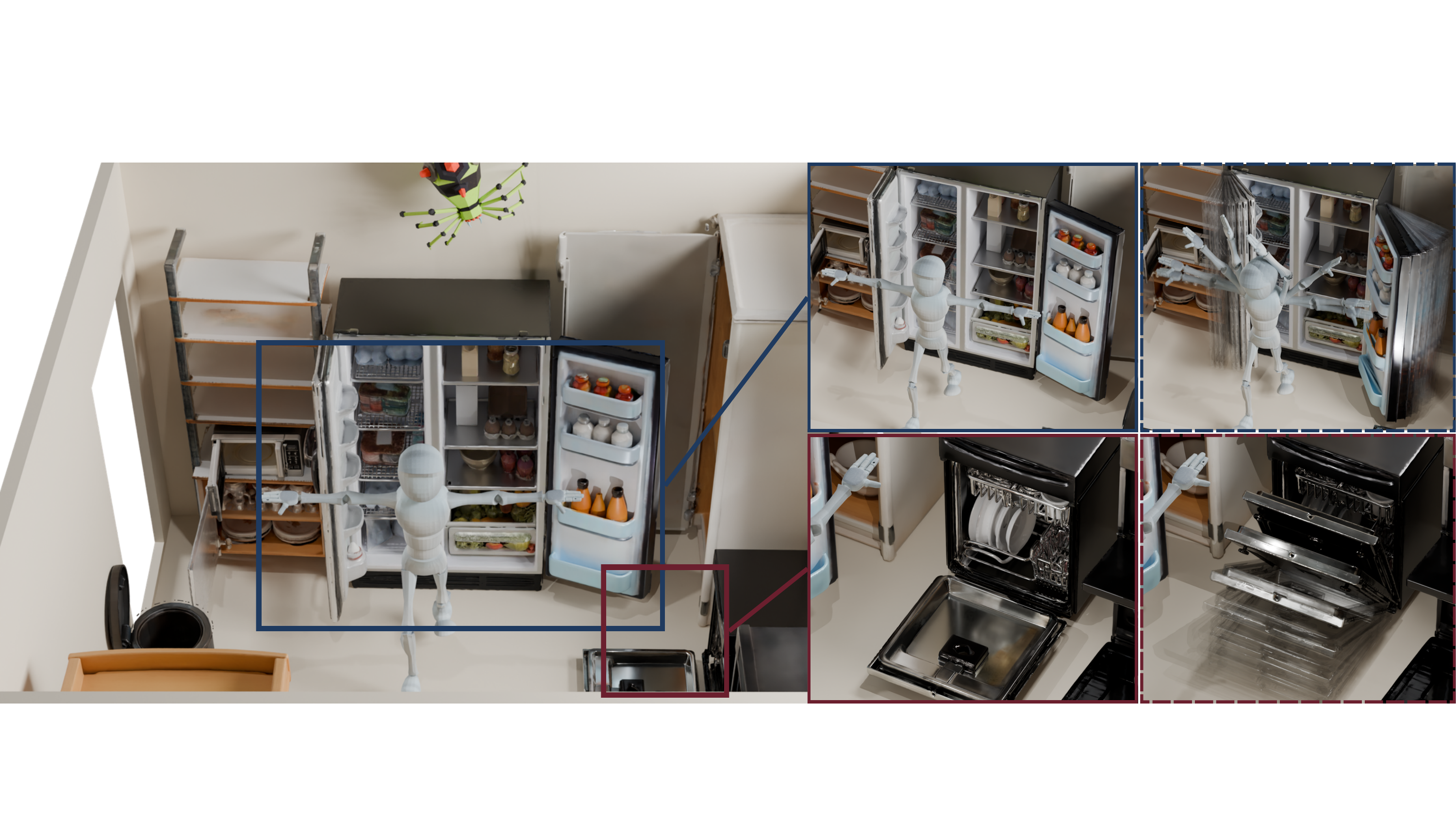}
  \captionof{figure}{\textbf{UnfoldArt} generates articulated 3D objects from a single text or image input, recovering both the external part structure and the high-fidelity interior geometry that becomes visible only under articulation. To achieve this, we propose a debate-driven agentic approach to articulated 3D reconstruction, combining hierarchical agent debate with a video generative prior that grounds articulation reasoning in concrete motion and exposes occluded interior geometry.}
  \label{fig:teaser}
\end{center}
\vspace{1em}

\begin{abstract}
Articulated 3D objects are essential for interactive environments in embodied AI,
robotics, and virtual reality, but reconstructing their structure and motion from
sparse observations remains challenging. Existing approaches remain largely
constrained by lack of supervised data or lack the priors needed to reliably
recover articulation, hidden geometry, and internal object structure.
We present the first debate-driven agentic approach to articulated
3D object reconstruction from text or image inputs that both grounds
articulation reasoning in concrete motion and exposes the occluded geometry
revealed under articulation.
High-level agents reason about object semantics and motion using knowledge from
vision-language and video models, while low-level agents estimate
articulation parameters and interaction points; together, they engage in a
two-round structured debate that first exploits global--local disagreement and then
grounds the agents in freely generated video.
The same video prior, conditioned on the agreed articulation, then drives each part through its motion to expose occluded
interiors and geometry that cannot be inferred from a single static view.
By combining agentic reasoning with a video generative prior, our approach
jointly infers articulation and reconstructs complete 3D articulated objects,
producing high-fidelity geometry, internal structure, and motion-consistent
states beyond directly observed surfaces.
\end{abstract}

\section{Introduction}

Articulated 3D objects form basic building blocks in interactive virtual worlds, underpinning applications across embodied AI, robotic simulation, and virtual reality. 
Generating them from text or image inputs would enable the creation of complex, manipulable scenes in which each object exposes not only its external form but also the hidden interior structures that become visible as its components are articulated. 
Realizing this requires jointly recovering an object's part decomposition, articulation parameters, and the occluded structure revealed through motion.

Due to their important role in enabling interactivity, significant efforts have recently been devoted to modeling articulated 3D objects. 
Most efforts have focused on training supervised models \cite{cao2025physx, liu2025singapo} directly on articulated 3D datasets such as PartNet-Mobility \cite{Xiang_2020_SAPIEN, Mo_2019_CVPR,chang2015shapenet}; this achieves strong in-domain results but struggles to generalize broadly, as available ground-truth data remains small in scale (on the order of 2k objects). Existing ground-truth data also typically models only exterior structure, omitting interior components, which further limits applicability. 

A more recent line of work reduces this reliance on articulated supervision by exploiting pre-trained generative priors~\cite{chen2025freeart3d, le2025articulate, lu2025dreamart}, yet each such method depends on a signal unavailable in the single-image, single-state setting we target: multi-view observations across articulation states~\cite{chen2025freeart3d}, a critic grounded in a fixed asset library with exact reference geometry~\cite{le2025articulate}, or a video prior fine-tuned on articulation-specific data~\cite{lu2025dreamart}. We defer a detailed discussion to Sec.~\ref{sec:related}; the key observation is that the mechanisms making these approaches reliable, namely extra views, a trusted critic, and a fine-tuned prior, are each absent in our setting.

Our approach addresses these challenges by recognizing that modern vision-language models (VLMs) and video generative models already encode rich knowledge about how objects articulate, and can be orchestrated in an agentic framing that deliberates over their predictions to recover robust, complete articulated 3D object structures. In particular, VLMs can supply high-level semantic reasoning about parts and plausible motion families, while video models can realize how these motions actually unfold, producing generated videos that ground the debate in concrete visual evidence about articulation parameters while also revealing occluded geometry that is often unrecoverable from a static view alone. Crucially, only two signals in our pipeline are specific to articulation: local-global agreement among agents, which resolves articulation parameters, and a frozen video prior, which serves the pipeline twice, first as external motion evidence grounding the debate and then as the means to expose occluded geometry. Every other component we build on (the 3D, image, and video priors) is a general-purpose pretrained model carrying no articulation supervision.

We thus propose a debate-driven agentic approach for articulated 3D object reconstruction from text or image inputs. 
From the input prompt or image, we first obtain a TRELLIS \cite{xiang2025trellis2} mesh that provides an initial 3D estimate of the object, giving the agents a grounded representation to reason over. 
A hierarchy of agents then divides labor between global and local reasoning: a Decomposer identifies parts and their plausible motions, a Grounder translates this into a local segmentation strategy on the mesh, and an Articulator predicts joint type, axis, and pivot for each part. Because low-level predictions are the most error-prone, the Articulator's output triggers a two-round debate between the global and local agents that first exploits their disagreement and then grounds the proposed motions in freely generated video to resolve articulation ambiguity. 
Once the debate converges, the agreed-upon articulation conditions a guided video inpainting pass that exposes the hidden geometry behind each movable part, which we reconstruct into an interactable URDF with complete interior geometry, ready for downstream use.

Our contributions are:
\begin{itemize}
    \item The first debate-driven agentic approach to articulated 3D reconstruction; our approach turns whole-object and part-level disagreement into the corrective signal for joint parameters, grounded in freely generated video of the proposed motion.
    
    \item Recovery of complete articulated objects, including occluded interiors, via 3D latent inpainting driven by articulation-conditioned generated video.
\end{itemize}
\section{Related works}
\label{sec:related}

\textbf{3D shape and part generation.} Recent 3D generative models produce high-fidelity static geometry from text or image inputs~\cite{zhang20233dshape2vecset, zhang2024clay, zhao2025hunyuan3d, xiang2024structured}, leveraging large 3D repositories such as Objaverse~\cite{objaverse, objaverseXL}. A complementary line decomposes shapes into parts, either generatively~\cite{tang2026efficient, yang2025omnipart, lin2025partcrafter, yan2025x, ma2025p3} or through segmentation~\cite{yang2024sampart3d, liu2025partfield}. These methods target static surface geometry. We build on these foundations, specifically TRELLIS~\cite{xiang2024structured, xiang2025trellis2}, and extend the generative paradigm to articulation and to the occluded interior structure such geometry omits.

\textbf{Articulation prediction for existing geometry.} A first family predicts articulation parameters for geometry that is already given, either estimating joints from RGB video~\cite{qian2022understanding} or fine-tuning vision-language models~\cite{huang2024a3vlm} on articulated datasets such as PartNet-Mobility (${\sim}2$K objects). These approaches recover only kinematics for pre-existing parts; we instead jointly generate geometry, articulation, and interior structure from a single image or text prompt.

\textbf{Articulated reconstruction from observations.} A second family jointly recovers geometry, parts, and kinematics, but remains tied to costly input signals. Per-instance optimization approaches~\cite{liu2023paris, weng2024neural, mu2021sdf, song2024reacto, wu2026reartgs} fit NeRF, SDF, or Gaussian representations to multi-view captures of an object in two or more articulation states. Feed-forward methods~\cite{liu2024cage, liu2025singapo, chen2024urdformer, dai2024automated, qiu2025articulate, gao2025meshart, li2025particulate, cao2025physx} instead learn priors over articulated structure from supervised data such as PartNet-Mobility; these include generative variants such as PhysX-Anything~\cite{cao2025physx}, which fine-tunes a VLM to output simulation-ready assets, yet still draws its articulation knowledge from the same supervised corpora. Both lines are thus constrained by their inputs: optimization requires expensive multi-view, multi-state captures of each object, while feed-forward methods depend on articulated training data that is small in scale and narrow in category coverage.

\textbf{Generative priors for articulation.} Closest to our work, a third family exploits pre-trained generative priors to reduce reliance on articulated supervision, yet each depends on a signal that is unavailable in our setting. Articulate-Anything~\cite{le2025articulate} runs a frozen VLM in an actor-critic loop over parts retrieved from a fixed 3D asset library; its critic is reliable in large part because retrieved parts carry exact reference geometry, so any anomaly in the assembled motion isolates to the joint and can be corrected with confidence. ArtiCraft~\cite{zhou2026articraft} also takes an agentic route, but its coding agent composes assets programmatically from primitives via an SDK, approximating object geometry; we instead exploit video and 3D generative priors to recover high-fidelity geometry faithful to the input image, including rich interior structure. We have no reference geometry to check against: our only observations are the agents' local and global views and a noisy generated video, none of which is trustworthy enough for a single critic to act on. We therefore replace the critic with a global-local debate whose disagreement itself serves as the corrective signal, treating video as evidence to weigh rather than obey, and we generate novel geometry with recovered interiors instead of assembling library parts. Unlike multi-agent debate that converges homogeneous agents toward consensus~\cite{du2023improving}, our agents are asymmetric and each is reliable on a different aspect of the input image, with the global agent authoritative on whole-object structure and the local agent on fine-grained part cues, so their disagreement, rather than their consensus, drives refinement. To our knowledge, ours is the first agentic method to settle articulation through a debate between agents instead of a single trusted critic, and the first to use a frozen video prior as the deciding evidence in that debate. FreeArt3D~\cite{chen2025freeart3d} repurposes a static 3D diffusion prior~\cite{xiang2024structured} via per-instance score distillation, but still requires sparse multi-view inputs across articulation states; we instead synthesize the articulated states ourselves through guided video generation. DreamArt~\cite{lu2025dreamart} shares our use of video as a geometry cue, but requires fine-tuning a video diffusion model on articulation-specific data; we keep the model frozen and steer it with a mask-anchored control strategy, avoiding that data requirement while retaining the prior's general coverage. The mechanisms behind these methods, a trusted critic and a fine-tuned prior, each presume a signal we forgo; our aim is to predict joints and reconstruct parts from the two signals that remain available without articulation supervision: local-global agreement for articulation and a frozen video prior for recovering geometry.

\newpage
\section{Method}
\label{sec:method}

\begin{figure}[t]
  \centering
  \includegraphics[width=\textwidth, trim={0.15\textwidth} 0 {0.18\textwidth} 0, clip]{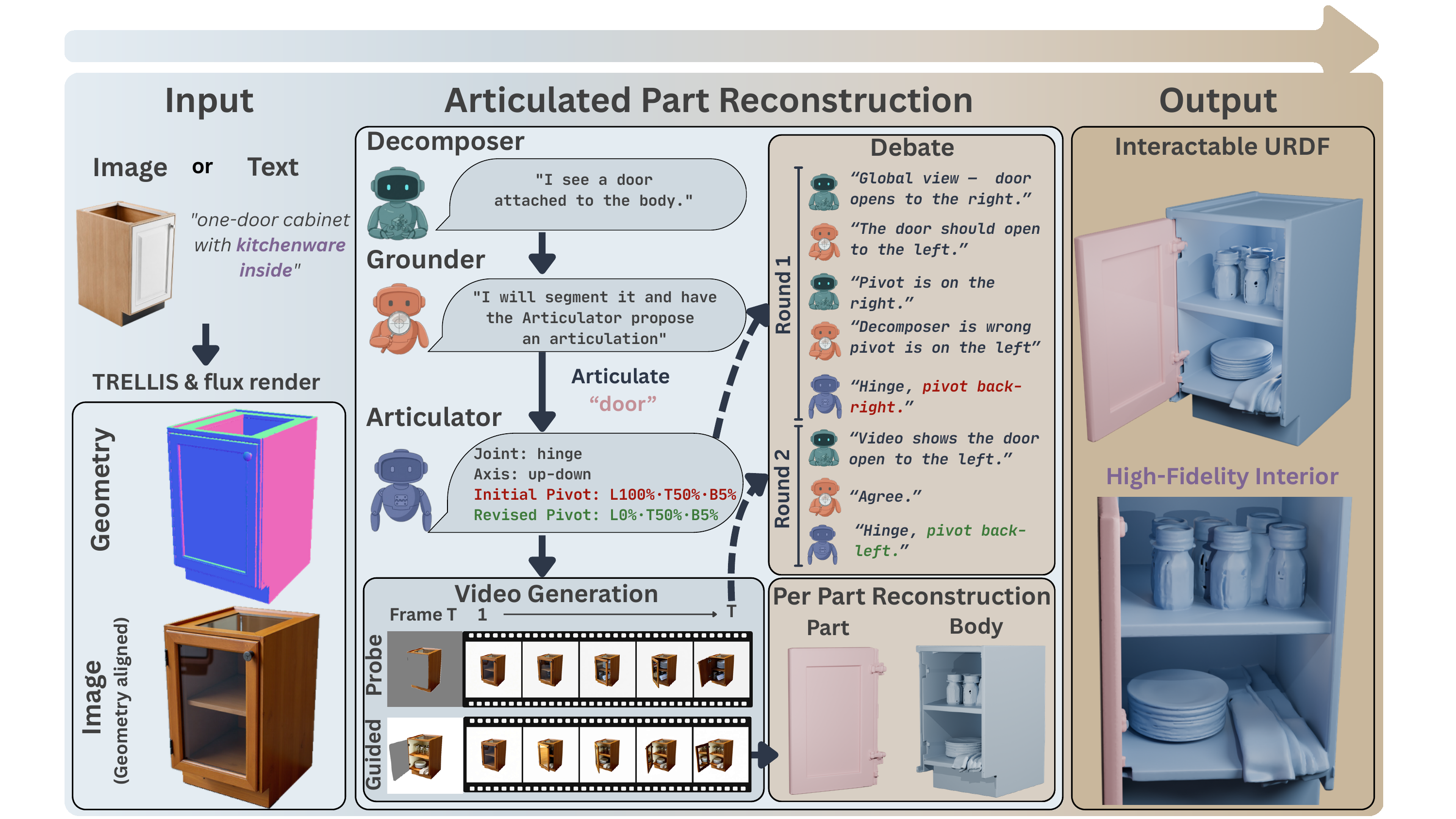}
\caption{\textbf{Method overview.} Given an unposed image or a text prompt as input, we generate a TRELLIS~\cite{xiang2025trellis2} mesh and a Flux~\cite{labs2025flux1kontextflowmatching, flux2024} render that provide a spatial and photorealistic visually-grounded representation for our agentic reasoning.
A hierarchy of three agents divides labor between global and local reasoning:
a \textit{Decomposer} reasons globally over parts and motion, a \textit{Grounder} chooses a segmentation strategy and orchestrates the per-part articulation order, and an \textit{Articulator} predicts the joint type, axis, and initial pivot. A two-round global-local debate then refines the prediction: the first round exploits whole-object–part disagreement, and the second grounds it in a freely generated video of the proposed motion, revising the pivot from initial (\textit{red}) to final (\textit{green}).
Conditioned on the articulation agreed upon in the debate, a guided video pass then drives each part through its motion, anchored to an end-frame mask representing the most articulated state. This also reveals any part interior, which drives per-part reconstruction into an interactable URDF with high-fidelity interior geometry.}
  \label{fig:method}
\end{figure}

Given an unposed image $y_i$ or text prompt $y_t$, we reconstruct a fully articulated 3D object with complete interior geometry, output as a URDF $\mathcal{U} = (\{M_i\}, \{J_i^\star\}, \mathcal{K})$ comprising part meshes, joints, and a kinematic tree. An overview of our approach is shown in Fig.~\ref{fig:method}.
Our design follows directly from the two signals available without articulation supervision. Because no single foundation-model prediction is trustworthy on its own, we resolve articulation through a structured global-local debate rather than a single query; and because a single static view cannot reveal occluded interiors, we use a frozen video prior both as motion evidence for that debate and as the means to expose the geometry hidden behind each movable part. The video prior thus serves two roles: a \emph{probe} video $V^{\text{probe}}_i$ as external evidence in the debate (Sec.~\ref{sec:debate}), and a \emph{guided} video $V^{\text{art}}_i$ that exposes each part's interior (Sec.~\ref{sec:video_inpainting}).

To give the agents a concrete object to reason over rather than text or a single view alone, we first reconstruct an initial 3D estimate of the object $M$ with TRELLIS~\cite{xiang2025trellis2} and re-render it through Canny-conditioned Flux~\cite{labs2025flux1kontextflowmatching, flux2024} to obtain a photorealistic image $I$ aligned with $M$, providing photorealistic, spatially-correlated input for VLM-based agents and for video generation. A hierarchy of LLM agents then divides labor between global semantic reasoning and local geometric prediction: movable parts are proposed, segmented in 2D, and an initial articulation $\tilde{J}_i$ is estimated for each part $i$, refined to $J^{(1)}_i$ in the first debate round and to the final $J_i^\star$ in the second. Finally, $J_i^\star$ conditions a guided per-part video pass whose final frame drives a 3D latent-inpainting stage that separates clean movable part and static body meshes.

\subsection{Hierarchical Agentic Reasoning}
\label{sec:agents}
We observe that articulation reasoning is most reliable when distributed across scales rather than collapsed into a single query.
Directly prompting an LLM to predict 3D articulation can yield unreliable results, as LLMs lack strong 3D understanding. We instead distribute the task across three agents at increasing levels of abstraction:
\begin{align}
\{(\ell_i, \tilde\tau_i)\} &= \mathcal{D}(I), &
m_i &= \mathcal{G}(I, \ell_i), &
\tilde{J}_i &= \mathcal{A}(I|_{m_i}, M|_{m_i}).
\end{align}
The \emph{Decomposer} $\mathcal{D}$ proposes part labels $\ell_i$ and coarse motion types $\tilde\tau_i$ from the image $I$. The \emph{Grounder} $\mathcal{G}$ converts the Decomposer's analysis into a 2D mask $m_i$ by prompting a segmentation model. This step is necessary because many parts are not text-addressable; for example, the chained links of a humanoid cannot be queried by name the way a \textit{door} or \textit{drawer} can. We therefore adapt the prompt to the object: text-recoverable parts are queried by class, while complex articulated objects are segmented with a generic \textit{``movable parts''} prompt. Our default backend is Gemini \cite{team2023gemini}, which reliably produces masks in both cases; SAM3 \cite{carion2025sam3segmentconcepts} is a stronger alternative on text-recoverable parts (Tab.~\ref{tab:ablation}). As a cheaper alternative we use a Flux2 variant, detailed in the appendix; since Flux2 cannot predict movable parts off the shelf, we prompt it for joints instead and use them for part segmentation.
Finally, the \emph{Articulator} $\mathcal{A}$ predicts the initial joint $\tilde{J}_i$ from a local crop $I|_{m_i}$ and the corresponding mesh region $M|_{m_i}$.

In the debate, the three agents take on distinct roles: $\mathcal{D}$ argues globally, $\mathcal{G}$ critiques locally, and $\mathcal{A}$ arbitrates. The Decomposer $\mathcal{D}$ argues from the full-object image and carries authority on whole-object structure, while the Grounder $\mathcal{G}$, which sees the full image during segmentation, is restricted in the debate to the local crop and its fine-grained cues such as hinges and seams. This global-local asymmetry between $\mathcal{D}$ and $\mathcal{G}$ is the foundation of our debate (Sec.~\ref{sec:debate}): when the two scopes disagree, the disagreement itself becomes an exploitable signal for refining the prediction. The Articulator $\mathcal{A}$ makes the initial local proposal $\tilde{J}_i$ and then steps out of the local role to serve as the arbiter between $\mathcal{D}$ and $\mathcal{G}$.
Agents reason in image coordinates, where the up-down, left-right, and back-front directions correspond to the $y$, $x$, $z$ axes of $M$.
Beyond these three reasoning agents, several auxiliary LLMs manage the generation flow (a \emph{Structurer} that assembles the kinematic tree $\mathcal{K}$, an \emph{Inpainter} that produces text prompts for WAN-VACE~\cite{vace} video generation, an \emph{Orientation} LLM that resolves each part's front-facing direction relative to the camera, and helpers for orchestration and frame quality gating); these are not central to our contributions and are detailed in the appendix.

\subsection{Articulation Estimation via Agent Debate}
\label{sec:debate}
We find that reliability, rather than capability, is often the bottleneck for foundation-model articulation estimation, and we therefore proceed through agentic deliberation. Each round grounds the prediction in progressively stronger evidence: the agents' own priors, then the global-local disagreement of Sec.~\ref{sec:agents}, and finally motion evidence external to the agents from video generation.
We start from the initial Articulator proposals $\tilde{J}_i = (\tau_i, a_i, p_i, [0, \theta_i^{\max}])$ for each part $i$, with joint type $\tau_i \in \{\text{revolute}, \text{prismatic}\}$, axis $a_i$, pivot $p_i$, and range $\theta_i^{\max}$. The Articulator does not emit these directly: it reports qualitative descriptors (rotation orientation, movement direction, pivot location, and interaction hardware), which a deterministic post-processing step maps to $(\tau_i, a_i, p_i)$. The range $\theta_i^{\max}$ is held at a fixed per-type default during the debate and recovered for the full object at URDF export.

\textbf{Round 1: global-local disagreement as evidence.}
An LLM prompted with the global image $I$ sometimes returns a different answer than when prompted with a local crop $I|_{m_i}$, and this asymmetry is itself informative: on a cabinet with two symmetric doors and no visible knobs, the global agent infers opposing motion from shape symmetry, whereas the local agent, lacking context, may guess wrong; when knobs mark the hinge side, the roles reverse. Disagreement between the two perspectives flags articulation ambiguity. Concretely, $\mathcal{G}$ flags any local contradiction (e.g., a handle on the proposed pivot edge, as in Fig.~\ref{fig:method}) and $\mathcal{D}$, as the whole-object authority, reconsiders its call in light of the flag; $\mathcal{A}$ then consolidates the exchange into $J_i^{(1)} = \mathcal{A}(\tilde{J}_i, c_\mathcal{D}, c_\mathcal{G})$, where $c_\mathcal{D}$ and $c_\mathcal{G}$ are the agents' critiques. The exchange is asymmetric: $\mathcal{G}$ may flag local evidence but cannot override $\mathcal{D}$ on whole-object identity (which door of a pair, which row of a stack), preventing a confident but unsupported consensus.

\textbf{Round 2: generated video as evidence.}
Round 1 consensus can still be wrong when priors and global-local cues agree on a confident but incorrect answer. We therefore add evidence external to the agents: a probe video $V^{\text{probe}}_i$ from a pure video prior~\cite{vace}, conditioned only on $I$ and generated once per part. A VLM quality agent inspects a five-frame strip of $V^{\text{probe}}_i$ and emits two signals for the debate: a quality judgment that gates the axis optimization of Sec.~\ref{sec:axis_refine}, and a motion prior describing how the part moves. Conditioned on this prior, $\mathcal{D}$ (full last frame) and $\mathcal{G}$ (crop) judge whether the part opened naturally and, when it did, propose a revised pivot from the observed swing rather than merely validating Round~1. A strict-consensus rule adopts the video-grounded $J_i^\star$ only when both sides find the pose plausible and agree on the pivot direction, otherwise falling back to $J_i^{(1)}$, discarding implausible generations while overriding $J_i^{(1)}$ when the motion is clear.

\subsection{Axis Refinement from Probe Video}
\label{sec:axis_refine}
For accepted generations, we track points sampled on $M|_{m_i}$ across $V^{\text{probe}}_i$ with optical flow~\cite{karaev24cotracker3, karaev23cotracker}, and fit a Rodrigues rotation (revolute) or shared translation (prismatic), minimizing
\begin{equation}
\mathcal{L} = \mathcal{L}_{\text{rep}}(\hat{y}, y) + \lambda_{\text{geo}}\, \mathcal{R}_{\text{geo}}(a, p, \hat{a}_{\text{LLM}}).
\end{equation}
Here $\mathcal{L}_{\text{rep}}$ is the Huber reprojection error and $\mathcal{R}_{\text{geo}}$ combines smoothness, monotonicity, and an agreement term pulling the fitted axis toward the quality agent's motion prior $\hat{a}_{\text{LLM}}$. Only the refined axis $\hat{a}$ returns to the debate; the pivot is left to the agents, as it is more sensitive to track noise.

\begin{figure}[t]
  \centering
  \includegraphics[width=\columnwidth]{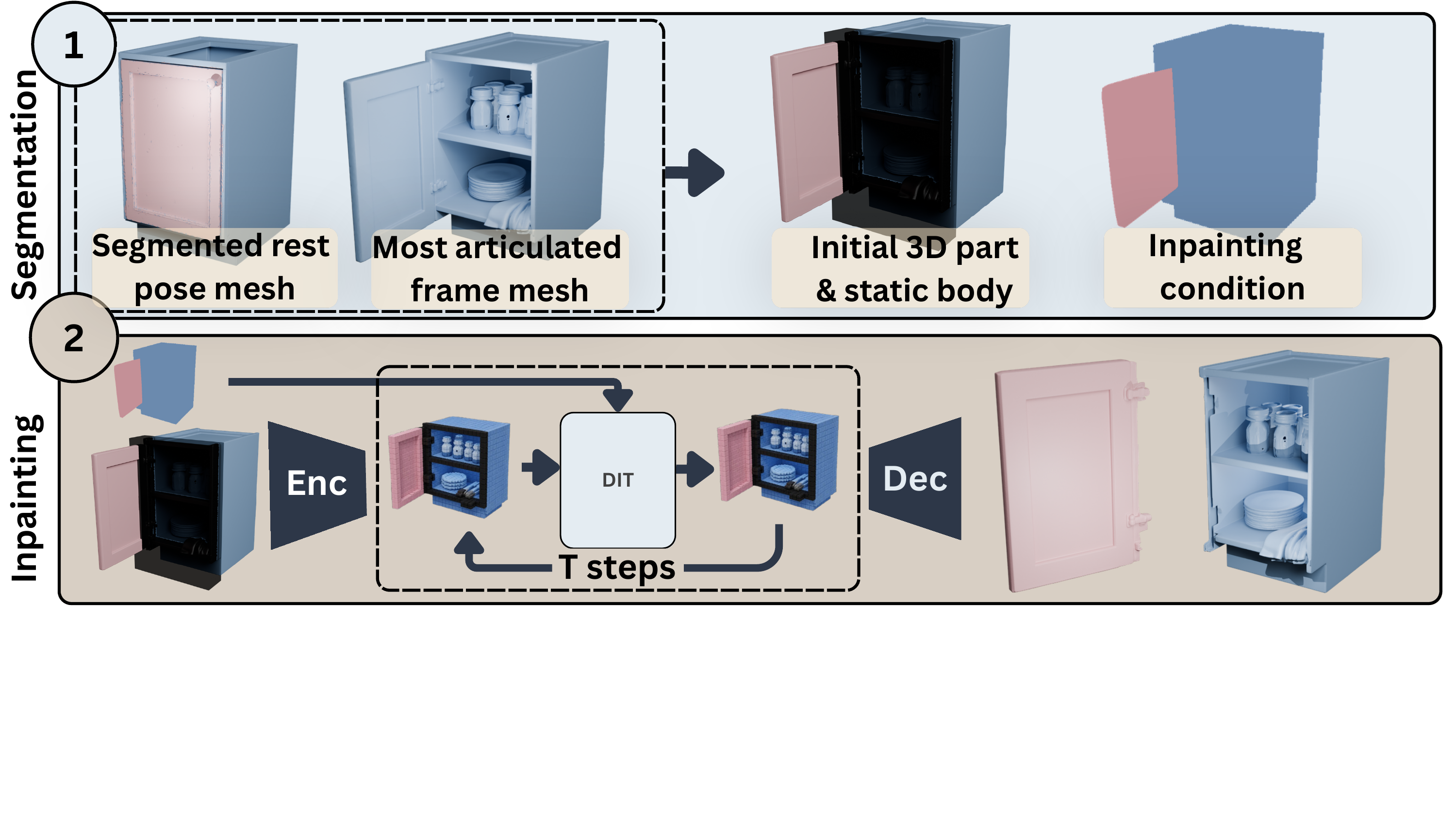}
  \vspace{-2.5cm}
\caption{
\textbf{Part reconstruction via 3D latent inpainting.}
The rest-pose mesh and mesh reconstructed from the maximum articulation frame in the guided video $V^{\text{art}}_i$ enable reconstruction of each part, here visualized for one part (pink).
Together, these meshes indicate part (pink) vs body (blue) components in an underlying voxel grid, considering the articulation sweep of the part. This provides an initial 3D segmented reconstruction, with only the voxels lying close to the boundary of part and body as ambiguous (black).
After rendering this partial assignment to 2D as the inpainting condition (right), we encode it into latents and denoise using RePaint to jointly resolve the part-body boundary and complete the previously occluded interior to yield clean separated meshes.}

  \label{fig:warped_mask_inpainting}
\end{figure}

\subsection{Articulation-Conditioned Video for Interior Recovery}
\label{sec:video_inpainting}
The agreed articulation $J_i^\star$ lets us recover the geometry occluded behind each movable part. We cast part-body separation as an inpainting problem in 3D latent space: rather than segmenting surface geometry directly, we commit the voxels we can label with certainty from motion and rest-pose evidence and inpaint only the contested boundary. This provides stronger constraints than surface-based segmentation such as SegViGen~\cite{li2026segvigenrepurposing3dgenerative}, which finetunes TRELLIS~\cite{xiang2025trellis2} to synthesize part colors on the surface and infers correspondence through color similarity; our construction instead derives a near-complete labeling from the part's articulation sweep and leaves only a thin ambiguous shell to resolve. Recovering this labeling requires a video that both drives the part through its motion and exposes the interior beneath it, which we synthesize next.

\textbf{Articulation-conditioned video.}
We synthesize a guided video $V^{\text{art}}_i$ explicitly conditioned on $J_i^\star$ that also reveals interior structures.
Starting from $I$, we remove all parts to obtain a body-revealing image $I^{\text{body}}$ using Flux, exposing previously hidden interior geometry such as cabinet cavities and drawer slots. For each part $i$ we transport its mesh by $J_i^\star$ and project to 2D, yielding a mask $m_i^{\text{open}}$ that represents the most articulated state and conditions the final video frame. Video generation then runs once per part, anchored on $m_i$ at the first frame and $m_i^{\text{open}}$ at the last, with intermediate frames free for the video model to fill in:
\begin{equation}
V^{\text{art}}_i = \text{WAN}(I^{\text{body}}, m_i, m_i^{\text{open}}), \quad \text{mask}(F_T) = m_i^{\text{open}},
\end{equation}
where $F_T$ is the final frame of $V^{\text{art}}_i$. The video model preserves object identity and motion consistency: WAN-VACE propagates the rest-pose part's texture, fine details (e.g., a door knob), and geometry through intermediate frames, so $F_T$ depicts the original part displaced along the predicted motion rather than a freshly synthesized object. In contrast, reasoning only through static image generation often results in identity drift. By keeping the body interior visible in every frame, this per-part animation provides the signal needed for articulated part segmentation and reconstruction. Since generated videos are not always usable, we vet the final frame before it drives reconstruction: an LLM scores $F_T$ for part integrity and motion plausibility, and generations where the part has drifted or disintegrated are rejected and re-run rather than reconstructed from a corrupted frame.

\textbf{Initial 3D part segmentation.}
$V^{\text{art}}_i$ provides strong visual signal as to both interior and part structures, which we lift to 3D. We reconstruct a TRELLIS mesh $M_i^{\text{open}}$ from the articulated final frame $F_T$ to obtain the geometry to be segmented, then label each of its voxels as part, body, or unknown. The joint trajectory $J_i^\star$ indicates where the part moves during articulation, and the body-revealing image $I^{\text{body}}$ indicates where the body sits at rest. Concretely, we lift the 2D part mask $m_i$ along $J_i^\star$ and sweep it from rest to maximum articulation, yielding the voxel volume $\Omega^{\text{sweep}}_i$ that the moving part traverses; we also lift $I^{\text{body}}$ into 3D to obtain $\Omega^{\text{body}}$. Voxels of $M_i^{\text{open}}$ in $\Omega^{\text{sweep}}_i$ belong to the part and voxels in $\Omega^{\text{body}}$ belong to the body, with ambiguity only along the boundaries of these sets. We mark this boundary shell as unknown, producing a binary mask $w$ with $w(j) = 1$ for committed voxels and $w(j) = 0$ for the shell, where $j$ indexes voxels of $M_i^{\text{open}}$.

\textbf{Resolving the boundary with latent inpainting.}
The committed part and body regions form a known latent $z_{t-1}^{\text{known}}$ paired with $w$, and we denoise the latent grid $z$ using RePaint~\cite{lugmayr2022repaint}:
\begin{equation}
z_{t-1} = w \odot z_{t-1}^{\text{known}} + (1 - w) \odot z_{t-1}^{\text{unknown}},
\end{equation}
where $z_{t-1}^{\text{known}}$ is the known voxels with noise added to match the current step, $z_{t-1}^{\text{unknown}}$ comes from the normal denoising step on $z_t$, and $\odot$ is the element-wise product. At each step, the part and body anchors are re-injected and information propagates into the unknown shell as the latent denoises. Once denoising completes, we group the voxels by similarity to the per-region anchor means in latent space and decode each group into a separate mesh; processing one part-body pair at a time reduces this grouping to a single binary partition. The resulting body and part meshes, together with $\{J_i^\star\}$ and $\mathcal{K}$, form the final URDF $\mathcal{U}$.
\begin{table}[t]
\centering
\caption{\textbf{Comparison with state of the art} on an Objaverse-animated OOD split (10 classes, 40 objects), an Objaverse-Household split (12 classes, 53 objects), and PartNet-Mobility (7 classes, 77 objects). RS/AS denote rest and articulated states.  Since FreeArt3D requires multi-state images, we supply additional inputs from our video-synthesis step. SINGAPO and PhysX-Anything are supervised on PartNet-Mobility. Side labels group methods as Sup.\ (supervised) and Zero-shot.}
\label{tab:main_comparison}
\resizebox{\textwidth}{!}{
\begin{tabular}{cl*{10}{c}}
\toprule

& \textbf{Method} &
\textbf{\makecell{RS-dgIoU\\($\downarrow$)}} &
\textbf{\makecell{AS-dgIoU\\($\downarrow$)}} &
\textbf{\makecell{RS-dcDist\\($\downarrow$)}} &
\textbf{\makecell{AS-dcDist\\($\downarrow$)}} &
\textbf{\makecell{RS-dCD\\($\downarrow$)}} &
\textbf{\makecell{AS-dCD\\($\downarrow$)}} &
\textbf{\makecell{AOR\\($\downarrow$)}} &
\textbf{\makecell{Axis Err\\($\downarrow$)}} &
\textbf{\makecell{Pivot Err\\($\downarrow$)}} &
\textbf{\makecell{Joint Acc\%\\($\uparrow$)}}  \\
\midrule
\multicolumn{12}{l}{\textbf{Objaverse OOD (10 classes)}} \\
\cmidrule(lr){1-12}
\multirow{2}{*}{\rotatebox[origin=c]{90}{\textbf{Sup.}}}& SINGAPO \cite{liu2025singapo}            & 1.049 & 1.049 & 0.171 & 0.197 & 0.154 & 0.201 & \underline{0.004} & \underline{53.5$^\circ$} & 0.306 & 37.6 \\
 & PhysX-Anything \cite{cao2025physx}     & 1.191 & 1.193 & 0.519 & 0.539 & 0.455 & 0.525 & 0.006 & 71.5$^\circ$ & 0.627 & 18.8 \\
\cmidrule(lr){2-12}
\multirow{3}{*}{\rotatebox[origin=c]{90}{\makecell{\textbf{Zero-}\\\textbf{shot}}}}& FreeArt3D \cite{chen2025freeart3d}           & \textbf{0.768} & \textbf{0.769} & \underline{0.156} & \underline{0.164} & \underline{0.133} & \underline{0.152} & \textbf{0.000} & 85.0$^\circ$ & 0.437 & 28.6 \\
& Articulate-Anything \cite{le2025articulate} & 1.387 & 1.388 & 0.242 & 0.264 & 0.158 & 0.227 & 0.011 & 67.0$^\circ$ & \underline{0.193} & \underline{47.8} \\
&  \textbf{Ours}       & \underline{1.044} & \underline{1.049} & \textbf{0.121} & \textbf{0.154} & \textbf{0.044} & \textbf{0.111} & 0.008 & \textbf{25.5$^\circ$} & \textbf{0.103} & \textbf{64.2} \\
\midrule
\multicolumn{11}{l}{\textbf{Objaverse-Household (12 classes)}} \\
\cmidrule(lr){1-12}
\multirow{2}{*}{\rotatebox[origin=c]{90}{\textbf{Sup.}}}
&SINGAPO \cite{liu2025singapo}            & 1.213  & 1.215  & 0.243  & \underline{0.360}  & 0.177 & 0.353 & \underline{0.005} & 39.8$^\circ$ & \underline{0.30} & 48.5 \\
& PhysX-Anything \cite{cao2025physx}     & 1.134  & 1.139  & 0.314  & 0.621  & 0.236 & 0.401 & 0.011 & 36.8$^\circ$ & 0.46 & 19.9 \\
\cmidrule(lr){2-12}
\multirow{3}{*}{\rotatebox[origin=c]{90}{\makecell{\textbf{Zero-}\\\textbf{shot}}}}
& FreeArt3D \cite{chen2025freeart3d}           & \textbf{0.926}     & \textbf{0.927}     & 0.183     & 0.406     & 0.248     & 0.388     & 0.13     & 25.0$^\circ$          & 0.34    & \textbf{58.4}   \\
& Articulate-Anything \cite{le2025articulate} & 1.178 & 1.183 & \underline{0.216} & 0.477 & \underline{0.136} & \underline{0.346} & \textbf{0.001}  & \underline{23.5$^\circ$} & 0.33 & \underline{57.5}   \\
& \textbf{Ours}       & \underline{0.955} & \underline{0.960} & \textbf{0.175} & \textbf{0.244} & \textbf{0.103} & \textbf{0.188} & 0.024 & \textbf{17.3$^\circ$} & \textbf{0.17} & \underline{57.5} \\
\midrule
\multicolumn{11}{l}{\textbf{PartNet-Mobility (7 classes)} }  \\
\cmidrule(lr){1-12}
\multirow{2}{*}{\rotatebox[origin=c]{90}{\textbf{Sup.}}}
& SINGAPO \cite{liu2025singapo}            & \textbf{0.570}  & \textbf{0.579}  & \textbf{0.076}  & \textbf{0.119}  & \textbf{0.028} & \textbf{0.049} & \textbf{0.001} & \textbf{1.4$^\circ$} & \textbf{0.022} & \underline{50.6} \\
& PhysX-Anything \cite{cao2025physx}     & 0.953  & 0.958  & 0.177  & 0.281  & 0.089  & 0.195  & \underline{0.006} & 21.6$^\circ$ & 0.193 & 34.7 \\
\cmidrule(lr){2-12}
\multirow{3}{*}{\rotatebox[origin=c]{90}{\makecell{\textbf{Zero-}\\\textbf{shot}}}}
&FreeArt3D \cite{chen2025freeart3d}           & 0.818  & 0.820  & 0.127  & 0.187  & 0.315  & 0.360  & 0.018  & 67.5$^\circ$ & 0.41 & 33.3 \\
& Articulate-Anything \cite{le2025articulate}& 0.949 & 0.953 & 0.145 & 0.207 & 0.057 & 0.115 & \textbf{0.001}  & \underline{7.5$^\circ$} & 0.36 & 72.3   \\
& \textbf{Ours}        & \underline{0.655}& \underline{0.662} & \underline{0.103} & \underline{0.154} & \underline{0.032} & \underline{0.078} & \underline{0.006} & 8.0$^\circ$ & \underline{0.15} & \textbf{77.5} \\
\bottomrule
\end{tabular}%
}
\end{table}

\section{Results}
\label{sec:experiments}

\subsection{Experimental Setup}
\label{sec:setup}
\paragraph{Datasets.}
We evaluate against different methods over 3D datasets with increasing difficulty. Across all three sets the joint types remain revolute or prismatic, and what changes is the object category and the structural complexity of part motion. First, PartNet-Mobility~\cite{Xiang_2020_SAPIEN,Mo_2019_CVPR,chang2015shapenet}, using the Singapo~\cite{liu2025singapo} test split (7 classes, 77 objects), in-domain for the supervised baselines. Second, an Objaverse-Household split from Objaverse-animated~\cite{objaverse, objaverseXL} (12 classes, 53 objects), where 4 categories overlap with the baselines' training set but carry more complex geometry and higher part counts, and the remaining 8 cover household categories with part motions absent from that distribution (e.g., back-hinged chest lids, pedal-actuated bins). Third, an Objaverse-OOD split from Objaverse-animated (10 classes, 40 objects) whose categories lie far outside curated articulation datasets, including helicopters, robotic arms, and humanoid figures, many of which also exhibit chained kinematics that retrieval- and template-based methods cannot represent. Full category lists are in the appendix.

\paragraph{Baselines.}
We compare with four state-of-the-art image-to-articulated-3D methods, spanning supervised and zero-shot regimes. Singapo~\cite{liu2025singapo} is a diffusion-based method trained on PartNet-Mobility that retrieves part meshes from a fixed database, the strongest retrieval-based baseline. PhysX-Anything~\cite{cao2025physx} fine-tunes a VLM for articulation prediction and uses TRELLIS~\cite{xiang2025trellis2} for part generation. Articulate-Anything~\cite{le2025articulate} is zero-shot, running a frozen VLM in an actor-critic loop that assembles part meshes from the PartNet-Mobility asset library; we run it from the same single image as our method. FreeArt3D~\cite{chen2025freeart3d} is also zero-shot, optimizing static 3D diffusion priors via SDS, but requires multi-view observations across articulation states, with no single-image mode; we supply them from frames of our guided WAN-VACE video, a favorable multi-state input rather than a degraded fallback.

\paragraph{Metrics.}
\begin{wraptable}{r}{0.5\textwidth}
\centering
\footnotesize
\setlength{\tabcolsep}{3pt}
\renewcommand{\arraystretch}{1.1}
\captionof{table}{\textbf{Ablations} (PartNet-Mobility). \textbf{(A)} joint prediction; the bottom row is the full configuration used in Tab.~\ref{tab:main_comparison}. \textbf{(B)} part reconstruction.}
\label{tab:ablation}
\resizebox{\linewidth}{!}{%
\begin{tabular}{l ccc}
\toprule
\multicolumn{4}{l}{\textbf{(A) Joint prediction}}\\
\midrule
\textbf{Variant} &
\makecell{\textbf{Axis Err}\\$(^\circ)\,\downarrow$} &
\makecell{\textbf{Pivot Err}\\$\downarrow$} &
\makecell{\textbf{Joint Acc}\\$(\%)\,\uparrow$} \\
\midrule
\multicolumn{4}{l}{\textit{Debate rounds (Llama 4, SAM3 segmentation)}}\\
Decomposer only (global)   & 9.29  & 0.255 & 79.6 \\
Grounder only (local)      & 15.71 & 0.244 & 77.5 \\
+ Round 1 (disagreement)   & 9.44  & 0.206 & 79.6\\
+ Round 2 (video evidence) & 8.27  & 0.206 & 79.6 \\
\addlinespace
\multicolumn{4}{l}{\textit{Segmentation backend (Llama 4, full debate)}}\\
SAM3            & 8.27  & 0.206 & 79.6 \\
Flux 2 Klein 9B & 18.00 & 0.225 & 58.8 \\
Flux 2          & 14.52 & 0.278 & 59.3 \\
Gemini          & 9.84  & 0.236 & 77.1 \\
\addlinespace
\multicolumn{4}{l}{\textit{Full config (Sonnet 4.6, Gemini, debate)}}\\
Ours (full) & 8.0 & 0.15 & 77.5 \\
\bottomrule
\end{tabular}%
}
\vspace{0.5em}
\resizebox{\linewidth}{!}{%
\begin{tabular}{l ccc}
\toprule
\multicolumn{4}{l}{\textbf{(B) Part reconstruction}}\\
\midrule
\textbf{Variant} &
\textbf{RS-dgIoU}\,$\downarrow$ &
\textbf{RS-dcDist}\,$\downarrow$ &
\textbf{RS-dCD}\,$\downarrow$ \\
\midrule
Probe video + SegViGen~\cite{li2026segvigenrepurposing3dgenerative} & 0.922 & 0.166 & 0.06 \\
\textbf{Guided video + inpainting (Ours)}   & \textbf{0.655} & \textbf{0.103} & \textbf{0.03} \\
\bottomrule
\end{tabular}%
}
\vspace{-3.5em}
\end{wraptable}
We adopt evaluation metrics from Singapo~\cite{liu2025singapo} and FreeArt3D~\cite{chen2025freeart3d}, covering geometry, articulation, and structural plausibility.
For geometry, we report Singapo's part-level distances over Hungarian-matched parts: bounding-box gIoU error (dgIoU), part-centroid distance (dcDist), and per-part Chamfer distance (dCD), all lower-is-better, in the rest state (RS-) and averaged across articulated states (AS-). We also report Singapo's average overlapping ratio (AOR), which detects unrealistic sibling-part collisions. For articulation, following FreeArt3D, we report joint-axis direction error (in degrees) and joint-pivot error (axis-to-axis distance for revolute joints). Since FreeArt3D assumes a known joint type, we additionally report joint-type accuracy (Joint Acc).
\paragraph{Implementation.}
All experiments are run on one NVIDIA A100 GPU. We use Claude Sonnet 4.6 for our reasoning agents and video quality screener (Llama was unreliable here), and Gemini for segmentation. Although SAM3 scores higher on PartNet-Mobility (Tab.~\ref{tab:ablation}), it is limited to text-recoverable parts; we default to Gemini because it also masks the chained, non-text-addressable parts in our Objaverse splits, where most of our evaluation lies. We report both backends in Tab.~\ref{tab:ablation}.
We generate 41 frames with WAN-VACE, set $\lambda_{\text{geo}} = 0.2$, and run the debate for two iterations.

\begin{figure*}[t]
  \centering
  \includegraphics[width=\textwidth]{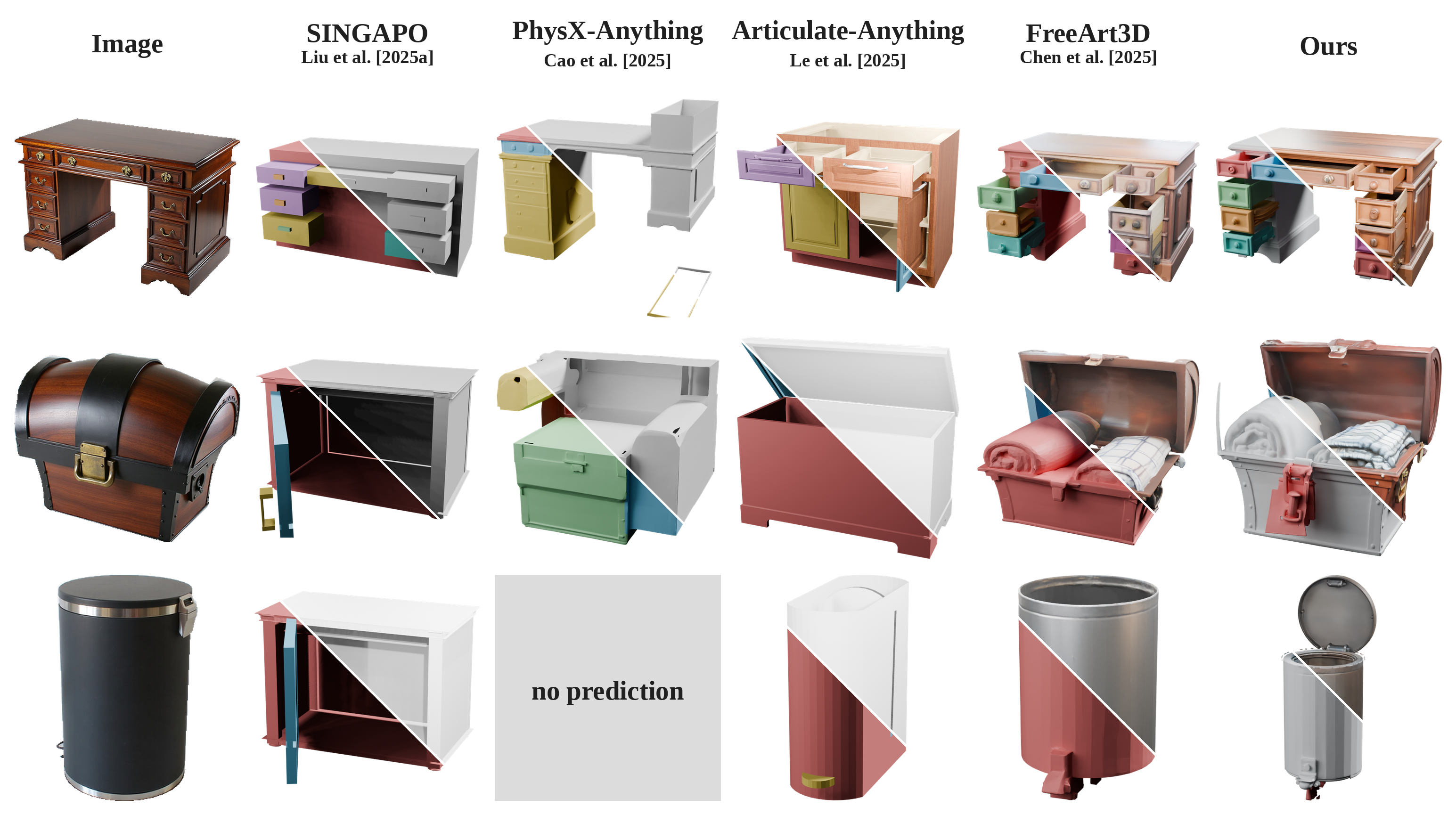}\\[1mm]

  \caption{\textbf{Qualitative results on Objaverse.}
  On objects outside curated articulation datasets, supervised baselines (PhysX-Anything, SINGAPO) struggle to align with the input geometry, while FreeArt3D recovers plausible structures but lacks fine-scale details and can struggle with segmentation. Our method recovers faithful per-part geometry and plausible articulation across diverse categories.
  }
  \label{fig:baseline_comparison_objaverse}
\end{figure*}

\begin{figure*}[t]
  \centering

  \includegraphics[width=\textwidth]{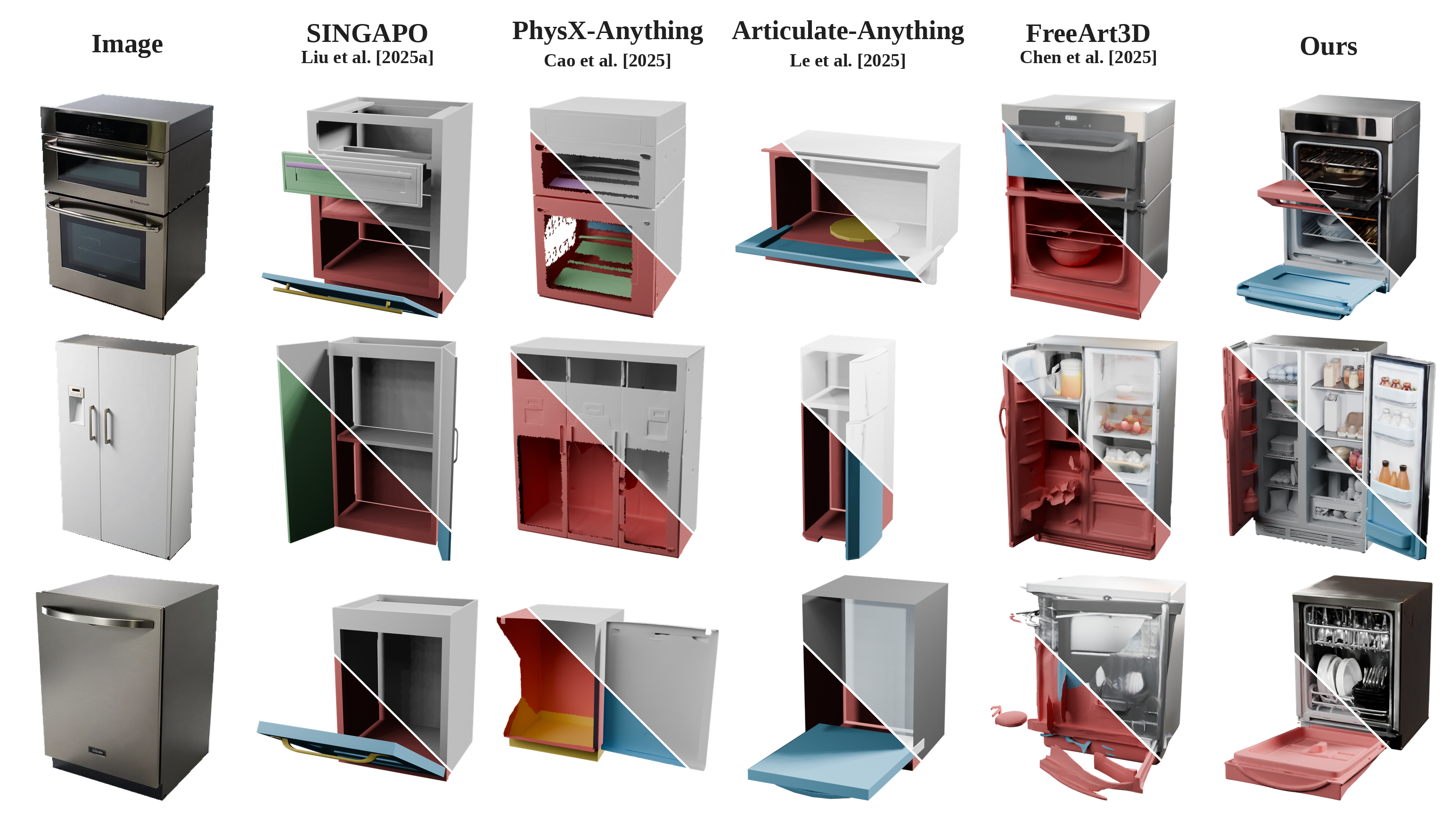}

  \caption{\textbf{Qualitative results on Partnet-Mobility.} 
  Our method remains on par with supervised baselines (Singapo, PhysX-Anything) performance on their in-domain data distribution, and produces more detailed and cleaner articulated reconstructions than the training-free FreeArt3D.
  }
  \label{fig:baseline_comparison_partnet}
\end{figure*}

\subsection{Comparison with State of the Art}
\label{sec:comparison}
\begin{figure*}[h]
  \centering
  \includegraphics[width=\textwidth, trim={0.55\textwidth} {0.5\textheight} {0.55\textwidth} 0, clip]{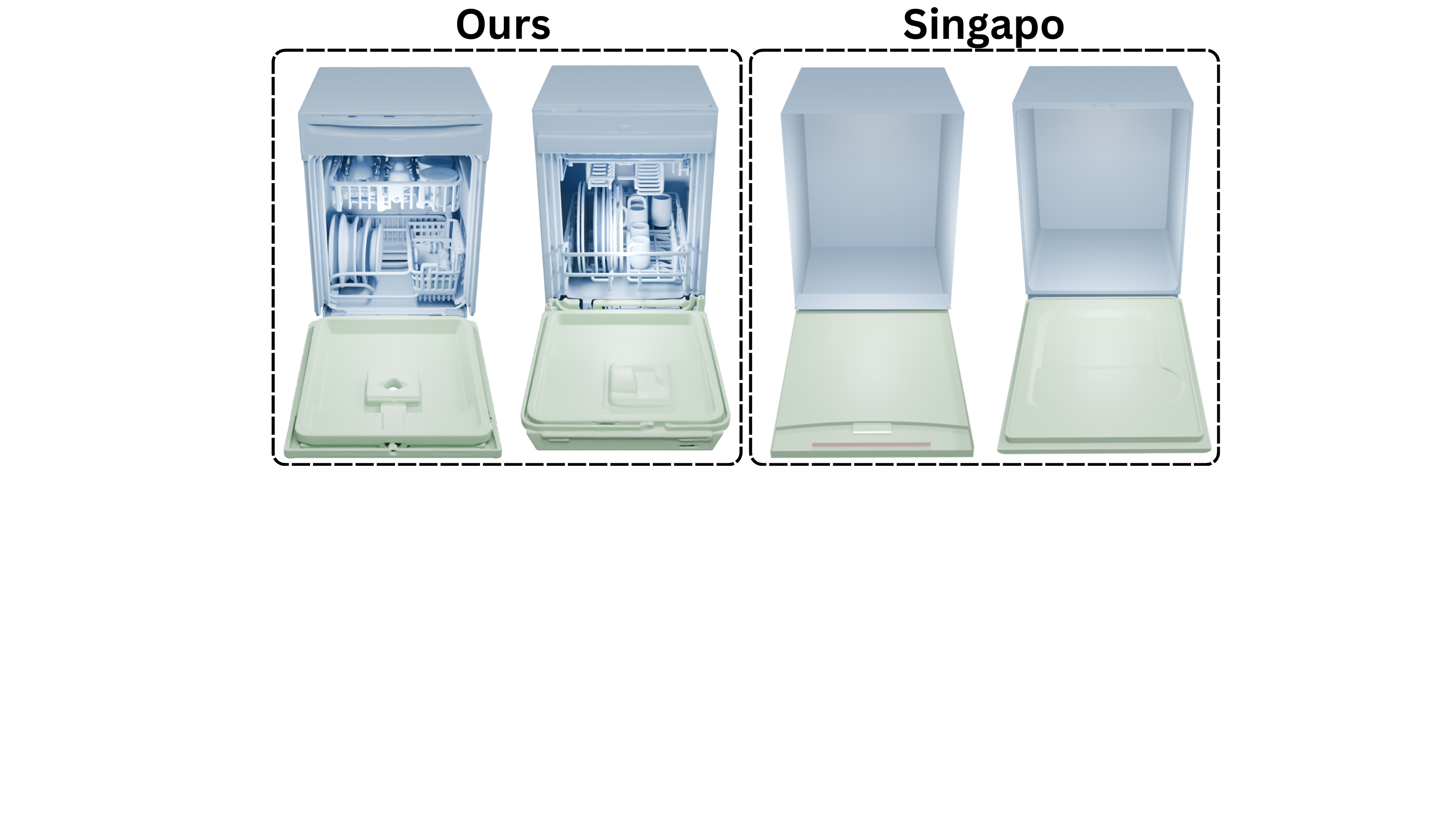}
  \caption{\textbf{Diversity of interior geometry.}
  Our approach recovers plausible, detailed,} category-consistent interior geometry hidden from view in the input image.
  \label{fig:interior_diversity}
\end{figure*}

\paragraph{Quantitative results.}
Table~\ref{tab:main_comparison} reports performance across the three sets. On the Objaverse-OOD split the supervised category priors do not apply, and although the joint types are the same revolute and prismatic ones seen in training, the object categories and in many cases the chained kinematics of robot arms and humanoid figures are novel, so both supervised methods degrade because retrieval and category templates cannot represent them. Our method is best here on the part-distance reconstruction and articulation metrics (dcDist, dCD, axis, and pivot error), sharply reducing axis error and leading on joint-type accuracy. The Objaverse-Household split behaves the same way on reconstruction and axis error; on joint-type accuracy we are on par with FreeArt3D there. Even in-domain on PartNet-Mobility, where Singapo performs best overall, our method stays close on reconstruction and reaches the highest joint-type accuracy of any method, despite never training on articulated data. FreeArt3D performs better over Objaverse-OOD and Objaverse-Household on dgIoU and AOR, where both metrics are sensitive to part count, since coarser decompositions yield fewer boxes to misalign and fewer siblings that can collide, independent of articulation quality. Its articulation is still poor, since it treats a single generated video as ground truth (Fig.~\ref{fig:baseline_comparison_objaverse}).
These results suggest our agentic approach generalizes across object categories and motion structures where supervised methods struggle, while remaining competitive in domain.
\paragraph{Qualitative results.}
Figures~\ref{fig:baseline_comparison_objaverse} and~\ref{fig:baseline_comparison_partnet} compare with state of the art on Objaverse and PartNet-Mobility. On the Objaverse-OOD objects, the supervised baselines struggle: PhysX-Anything defaults to a generic cabinet shape that ignores the input geometry, while SINGAPO produces coarse bounding-box articulation that loses object identity. FreeArt3D aligns more faithfully with the input but its SDS optimization limits fine-scale detail and can fail to segment parts. Our method recovers faithful per-part geometry and plausible joints across all examples. On PartNet-Mobility, we remain on par with the supervised methods while producing visibly cleaner articulated states, opening parts fully to reveal detailed internal components; Fig.~\ref{fig:interior_diversity} (appendix) further highlights plausible, category-consistent interior geometry such as racks and shelving.
\subsection{Ablation Study}
\label{sec:ablations}
\paragraph{Agent debate vs. single-agent prompting.}
Tab.~\ref{tab:ablation}(A) compares single-agent prompting against our debate, with the reconstruction stage fixed and Llama~4 as backbone so the debate is isolated from VLM strength; the final Sonnet row confirms the same ordering. The two single-agent variants fail in complementary ways: the global view alone (Decomposer) predicts axes reasonably but mislocates pivots, while the local crop alone (Grounder) is worst on axes. Round~1 arbitrates this disagreement and recovers the strengths of both, cutting pivot error well below either single agent while holding axis error at the global agent's level.
Joint-type accuracy stays consistent across settings, as type is a coarse categorical choice a single VLM call resolves reliably; small differences across backbones reflect this rather than any regression. The debate targets the geometric parameters where global and local cues can disagree. We also show that the video evidence further improves the performance.
\paragraph{Does articulation-guided reconstruction improve over off-the-shelf alternatives?}
Tab.~\ref{tab:ablation}(B) ablates our reconstruction stage. Guided video and 3D latent inpainting share one construct: the part's articulation sweep from $J_i^\star$ anchors the end-frame mask that conditions video generation (Sec.~\ref{sec:video_inpainting}) and commits the voxel labels that drive inpainting. We ablate both at once, swapping the full design for free (unconditioned) WAN-VACE \cite{vace} video and SegViGen~\cite{li2026segvigenrepurposing3dgenerative} surface segmentation. Every rest-state metric drops substantially: without the sweep the video can leave the interior occluded, and surface segmentation has to read the part-body boundary from color alone rather than from committed motion. Grounding reconstruction in the agreed articulation is what recovers the interior faithfully.

\paragraph{Limitations.}
Despite the flexibility of our approach, various limitations remain. 
Our agentic framing employs several VLM queries as well as two video generations per part, trading generality and robustness for runtime speed; a full reconstruction ranges from approximately 13 minutes to 1 h per object when the Orchestrator executes the full pipeline. 
Our approach also focuses on parts that move, and so does not further decompose static regions into their finer-scale semantic components. 
In addition, very small or low-amplitude articulations, such as flat buttons, are difficult to capture through a video prior, since the corresponding motion is too subtle to manifest reliably in generated video.
\section{Conclusion}
We presented the first debate-driven agentic approach to articulated 3D object
reconstruction from text or image inputs, resolving articulation ambiguity
through a structured global-local debate that exploits
disagreement between whole-object and part-level views and adjudicates it
against freely generated video, rather than relying on a
trusted critic or articulation supervision. Our key insight is that
vision-language and video generative models already encode rich knowledge
about how objects articulate, and that orchestrating them as deliberating
agents surfaces this knowledge reliably. A guided video inpainting pass
conditioned on the agreed articulation then exposes the
geometry occluded behind each movable part, yielding interactable URDFs with
both faithful articulation and complete interior structures. Our method
recovers articulated objects far beyond the object categories and
motion structures available for supervised training, from household objects with
unseen part motions to chained kinematics such as robotic arms, and recovers
interior structures that prior methods do not address, opening new
opportunities for articulated object synthesis beyond static priors.

\section*{Acknowledgements}
This work was supported by the ERC Starting Grant SpatialSem (101076253).

\medskip
{
\small
\bibliographystyle{plainnat}  
\bibliography{main}     
}
\newpage
\section{Appendix}

\subsection{Evaluation Categories}
\label{app:categories}
Tables~\ref{tab:cats_furniture} and~\ref{tab:cats_ood} list the categories and per-category object counts of our two Objaverse-animated evaluation splits (Sec.~\ref{sec:setup}).
\begin{table}[h]
\centering
\footnotesize
\caption{\textbf{Objaverse-Household split} (12 classes, 53 objects). The first 4 categories overlap with the supervised baselines' training categories; the remaining 8 exhibit part motions absent from their training distribution.}
\label{tab:cats_furniture}
\begin{tabular}{lc|lc}
\toprule
\textbf{Category} & \textbf{\#Obj} & \textbf{Category} & \textbf{\#Obj} \\
\midrule
\multicolumn{4}{l}{\textit{Overlapping with PartNet-Mobility}}\\
Refrigerator & 4 & Microwave oven & 3 \\
Desk / table & 4 & Dresser / drawers & 8 \\
\midrule
\multicolumn{4}{l}{\textit{Unseen part motions}}\\
Treasure chest & 11 & Stove / range & 3 \\
Box / crate / case & 6 & Door & 3 \\
Cabinet / wardrobe & 6 & Small appliance & 2 \\
Trash bin & 2 & Faucet & 1 \\
\bottomrule
\end{tabular}
\end{table}
\begin{table}[h]
\centering
\footnotesize
\caption{\textbf{Objaverse-OOD split} (10 classes, 40 objects), featuring object categories far outside curated articulation datasets, many of which also show chained kinematics.}
\label{tab:cats_ood}
\begin{tabular}{lc|lc}
\toprule
\textbf{Category} & \textbf{\#Obj} & \textbf{Category} & \textbf{\#Obj} \\
\midrule
Helicopter & 7 & Multi-legged creature & 4 \\
Aircraft / spacecraft & 3 & Sci-fi pod & 5 \\
Robotic arm & 4 & Wind turbine & 4 \\
Humanoid robot & 5 & Animal & 3 \\
Humanoid figure & 2 & Vehicle / machinery & 3 \\
\bottomrule
\end{tabular}
\end{table}
\subsection{Interior diversity}

\subsection{FLUX-based Joint Detection}
\label{app:flux_joints}

\begin{figure}[h]
  \centering
  \includegraphics[width=\columnwidth]{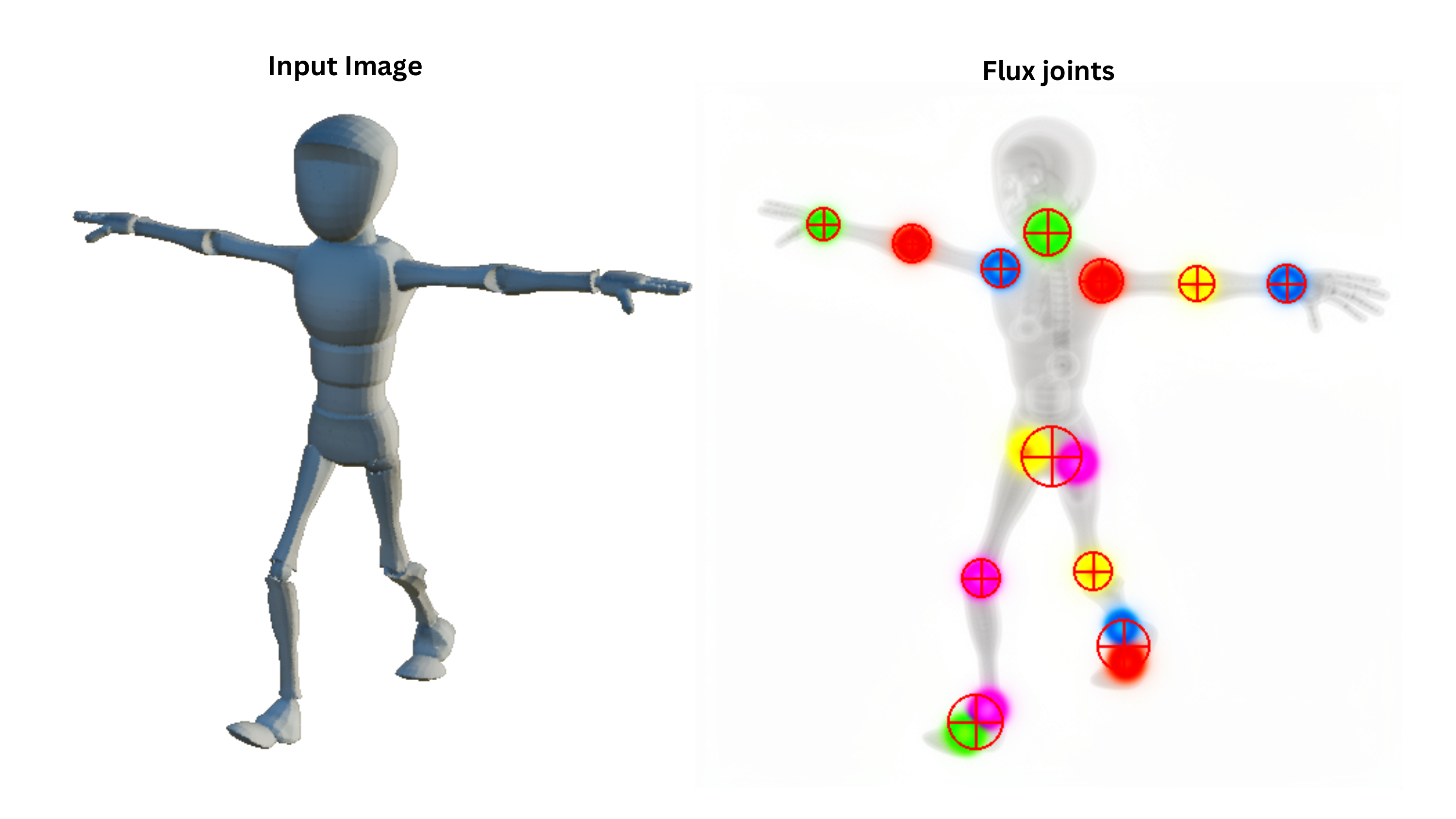}
  \caption{\textbf{Noisy part segmentation from FLUX joint detection.} FLUX generates images with high-saturation joint regions; connected components on the saturation maps are lifted to 3D to localize the joint. After removing the joint region, DBSCAN clustering on the remainder yields the noisy part segmentation used as input to our inpainting stage.}
  \label{fig:flux_joints}
\end{figure}

For parts that cannot be reliably localized via text queries, we obtain a noisy 3D segmentation by leveraging FLUX as a joint detector. The pipeline is illustrated in Figure~\ref{fig:flux_joints} and proceeds as follows.

\textbf{Joint detection via saturation.} We prompt FLUX to generate images in which candidate joint regions are rendered with high saturation. Running connected-component analysis on the resulting 2D saturation maps yields per-view joint proposals.

\textbf{Lifting to 3D.} The 2D joint regions are lifted into 3D by aggregating consistent components across views, producing a coarse 3D mask of the joint region.

\textbf{Part extraction.} We remove the joint region from the object and cluster the remaining surface points with DBSCAN. The resulting clusters serve as the noisy part segmentation that is subsequently refined by our inpainting stage.

\subsection{Agent Prompts}
\label{sec:supp_prompts}

This supplementary contains the system prompts used for each agent in our pipeline. Prompts are presented under the method subsection they implement. Output schemas and minor formatting rules are abbreviated for readability; full prompts are available in our code release. Prompts are reproduced verbatim as used in our experiments.

\subsection{Hierarchical agents (\S\ref{sec:agents})}

\paragraph{Decomposer (Stage 1).}
The Decomposer consumes the rendered image $I$ and proposes parts and coarse motion guesses, producing the initial part hypothesis used to drive segmentation. Its output is kind-level, not instance-level: ``three drawers'', not ``drawer\_0, drawer\_1, drawer\_2''.

\begin{promptbox}{Decomposer system prompt}
You are looking at a photograph of a 3D object. Your job is to give a short, high-level read of what kinds of parts this object has, roughly how each kind moves, and roughly how the kinds fit together.

This is a rough first pass. You are not labelling individual instances and you are not counting them precisely. Somebody else will look at the 3D mesh and pick out each specific instance later.

\medskip

\textbf{How to think about it.} Think in kinds, not instances. If you see three stacked drawers, you say ``drawers --- multiple, stacked vertically, they slide straight out toward the viewer''. You do not say ``drawer\_0, drawer\_1, drawer\_2''. Keep it rough --- you are pointing at categories and gestures, not measurements.

\medskip

\textbf{What you must produce.} A single object-level note with: \texttt{object\_type} (a short natural name --- ``wooden dresser'', ``ceramic teapot''); \texttt{parts} (a list of part kinds, each with \texttt{part\_type}, \texttt{description}, a \texttt{motion} block containing \texttt{moves}, \texttt{direction}, and \texttt{reasoning}, and a \texttt{count\_hint} like ``one'', ``multiple stacked vertically''); and a \texttt{hierarchy\_sketch} (one or two sentences on how the kinds fit together).

Do not pin down exact pivots. Do not mention axes, angles, degrees, or percentages --- that level of detail is for later steps.

\medskip

\textbf{Rules.} Plain everyday language only --- no X/Y/Z, no coordinates, no technical terms. Use ``left/right/forward/back/up/down'' --- nothing more technical. Do not mint ids like ``drawer\_0''. Do not invent kinds you cannot clearly see. If a kind does not move, set \texttt{moves} to false and \texttt{direction} to null.
\end{promptbox}
\newpage
\paragraph{Grounder (Stage 2).}
The Grounder receives the Decomposer's kind list and selects a segmentation route per kind: a text-query route for text-queryable parts (doors, drawers, handles), or a joint-driven approach for chained kinematics that text queries cannot resolve (humanoid figures, robot arms, spider legs). The body is never queried explicitly; it falls out as the complement of the part masks.
In the prompt below, ``SAM'' denotes the segmentation backend generically; our default backend is Gemini, with SAM3 as an alternative (Tab.~\ref{tab:ablation}).
\begin{promptbox}{Grounder system prompt}
You decide what SAM should look for in the input photograph. The Decomposer has already told you what kinds of parts the object has, roughly how many of each, and how each kind tends to move. Your job is to turn that into a list of plain kind-level phrases for SAM, plus a motion note per kind.

SAM runs once with all of your phrases together, returning one mask per instance --- if there are three drawers and you say \texttt{"drawer"}, SAM returns three drawer masks. The tool layer mints one part id per mask, ordered by position.

\medskip

\textbf{Writing good SAM phrases.} Use the bare kind name (\texttt{door}, \texttt{drawer}, \texttt{handle}, \texttt{knob}, \texttt{wheel}, \texttt{shelf}, \texttt{lid}). \emph{No positional qualifiers} --- never ``top drawer'', ``upper door'', ``left handle''. SAM does not understand positional words. Add a descriptor only when the class name is ambiguous (\texttt{cabinet door}, \texttt{fridge door}). The body always falls out as the complement of the part masks --- never list it as a SAM phrase.

\medskip

\textbf{When to use the joint-driven route instead of SAM.} The decision is about \emph{kinematic chaining}. Use SAM when every moving part pivots directly on the body --- doors, drawers, lids, wheels, knobs on cabinets, fridges, dressers, microwaves, suitcases. Use joints when a part's pivot lives on another moving part: humanoid figures (foot $\to$ shin $\to$ thigh $\to$ hip), articulated robot arms, spider/octopus/crab legs, quadrupeds, excavators, articulated lamps. SAM masks alone cannot recover per-link pivots in chains.

\medskip

\textbf{Output.} \texttt{sam\_kinds} (one entry per kind, with \texttt{sam\_prompt}, \texttt{part\_type}, \texttt{motion\_guess}), an optional \texttt{joints\_kinds} for chained kinds, a \texttt{body\_motion} note, and a \texttt{reasoning} field.

\medskip

\textbf{Rules.} Plain language only. Stay close to the Decomposer's note --- if it names kinds you cannot see in the photo, flag it in \texttt{reasoning} rather than silently dropping or inventing kinds. Do not refine motion --- copy from the Decomposer's note. A later step handles precise pivots and rotations.
\end{promptbox}
\newpage
\paragraph{Articulator (Stage 3).}
The Articulator predicts the initial joint $\tilde{J}_i$ from a local crop around each part. It reasons via a hardware-anchored chain: find the visible interaction point, place the pivot on the opposite edge, and sanity-check that the motion sweeps the handle toward the pivot. The Articulator's output is local-only: any whole-object identity question (which door of a pair, which row of a stack) is deferred to the Decomposer in the debate.

\begin{promptbox}{Articulator system prompt}
You are looking at a close-up crop of one part of a 3D object. Your job is to say where the pivot sits inside this crop --- the point the part hinges or pulls away from. The crop has direction labels drawn on its edges (left/right, top/bottom, FRONT/BACK).

\medskip

\textbf{Your scope --- local-crop detail only.} You are the LOCAL-DETAIL agent. Reason from what is directly visible in this crop: handle position, visible hinge gap or shadow, seam direction, how the red-outlined edges sit inside the body. A downstream WHOLE-OBJECT critic sees the full image and is responsible for any question that depends on the object as a whole (``which door of a pair'', ``which row of drawers''). You MUST NOT answer those questions. Anchor pivot calls on local evidence only.

\medskip

\textbf{Chain of reasoning --- anchored on the visible hardware.} Reason in this sequence and state each step explicitly:

\textit{Step 1 --- Find the hardware in the rest crop.} Identify the visible interaction point (handle / knob / pull / bar) and name which edge it sits on: \texttt{right}, \texttt{left}, \texttt{top}, \texttt{bottom}, or \texttt{front face / centred}.

\textit{Step 2 --- The pivot is on the OPPOSITE edge.} Apply the table:

\smallskip
\begin{center}\footnotesize
\begin{tabular}{ll}
\toprule
Hardware sits on... (closed) & Pivot edge $\to$ percentage \\
\midrule
\texttt{right} edge & \texttt{left} edge $\to$ \texttt{from\_left} $\approx$ 0\% \\
\texttt{left} edge & \texttt{right} edge $\to$ \texttt{from\_left} $\approx$ 100\% \\
\texttt{top} edge (oven door, drop-front) & \texttt{bottom} edge $\to$ \texttt{from\_top} $\approx$ 100\% \\
\texttt{bottom} edge (flip-up cabinet) & \texttt{top} edge $\to$ \texttt{from\_top} $\approx$ 0\% \\
\texttt{front} centred (slider) & no hinge --- all three $\approx$ 50\% \\
\texttt{front} edge (chest lid) & \texttt{back} edge $\to$ \texttt{from\_back} $\approx$ 0\% \\
\bottomrule
\end{tabular}
\end{center}
\smallskip

\textit{Step 3 --- Motion sanity-check.} When the part opens, the handle traces an arc and sweeps TOWARD the hinge. If \texttt{movement\_direction} and \texttt{pivot\_estimate} are incompatible (handle on right but \texttt{from\_left} $\approx$ 100\%), redo step 1.

\medskip

\textbf{Output.} \texttt{pivot\_estimate} (\texttt{from\_left}, \texttt{from\_top}, \texttt{from\_back} as percentages), \texttt{rotation\_orientation} ($\in$ vertical, horizontal, depth, none-slides), \texttt{movement\_direction}, \texttt{interaction\_hardware}, and \texttt{reasoning} that explicitly walks through steps 1$\to$2$\to$3.

\medskip

\textbf{Rules.} Plain everyday language. No X/Y/Z, no coordinates, no degree numbers. Always read FRONT/BACK labels --- a pivot near BACK signals a rear attachment.
\end{promptbox}
\newpage
\subsection{Round 1 debate: prior-based global/local critique (\S\ref{sec:debate})}

\paragraph{Decomposer (Round 1).}
The Decomposer critiques the Articulator's $\tilde{J}_i$ from the global perspective, with access to the full-object image and the Stage-1 kind-level read. It is the highest authority in the debate: the Grounder is a local validator and cannot override the Decomposer on whole-object identity.

\begin{promptbox}{Decomposer Round 1 system prompt}
You are the Decomposer in a structured debate about how one specific articulated part moves. \textbf{You are the HIGHEST authority in this debate.} The Grounder validates local crop-level details (visible hinge gap, handle position) but cannot override your whole-object call --- they only see a zoomed crop, no whole-object context, no kind-level prior.

You are shown: (1) the FULL object image with a RED CONTOUR around the debate-target part --- \emph{the Grounder does not see this image}, only the zoomed crop; (2) the part crop; (3) your own earlier prior; (4) the Articulator's proposal; (5) the Grounder's latest argument.

\medskip

\textbf{Your scope --- whole-object identity is yours alone.} You are the SOLE authority on: which door of a symmetric pair this is; which row/column of a grid of drawers; which side of the body the part sits on; what kind of object this is. The Grounder has been explicitly told not to answer these. You OWN these calls. Pin every \texttt{from\_left}/\texttt{from\_top}/\texttt{from\_back} percentage to a whole-object observation first (``in the full image the red contour is on the right door of a pair, therefore the hinge is on the right edge'').

\medskip

\textbf{Reasoning chain.} Apply the same hardware $\to$ pivot table as Stage 3, but anchored on the FULL-object image. Highlight at least one whole-object observation the Grounder cannot make from the crop alone (``looking at the full cabinet I can see this is the right door of a pair, so the hinge must run on the right edge''). Stand behind your prior unless the Grounder raises a concrete LOCAL visual point that reframes your read --- a Grounder claim about whole-object identity is not a legitimate challenge.

\medskip

\textbf{Output.} \texttt{motion\_position}, \texttt{pivot\_position}, \texttt{interaction\_position}, \texttt{response\_to\_articulator} (with concrete field-level edits and target values), and \texttt{response\_to\_other\_side}, all in full sentences.

\medskip

\textbf{Rules.} Plain everyday language. Do not converge out of politeness --- if kind-level convention says the Articulator is wrong, keep saying so. Do not propose 3D axes or rotation matrices.
\end{promptbox}
\newpage
\paragraph{Grounder (Round 1).}
The Grounder critiques $\tilde{J}_i$ from the local perspective, with access only to the zoomed part crop. Its role is strictly validation: it can flag concrete local contradictions (handle visible on the same edge as the proposed pivot) but cannot override the Decomposer on object identity or kind-level convention.

\begin{promptbox}{Grounder Round 1 system prompt}
You are the Grounder in a structured debate. \textbf{Your role is strictly local validation.} The Decomposer is the highest authority --- it has the full-object image AND the canonical Stage-1 kind-level read. You check whether the Decomposer's whole-object call is consistent with what the zoomed crop shows pixel-level. You are NOT here to override the Decomposer on motion / pivot / object-type / which-side-of-pair --- those are out of your scope.

You are shown: (1) the part crop with a RED CONTOUR; (2) your own earlier prior; (3) the Articulator's proposal; (4) the Decomposer's latest argument.

\medskip

\textbf{Validation behaviour.} If the Decomposer's pivot call places the hinge on edge X, check the crop: is the visible handle on the OPPOSITE edge? Is the visible hinge gap or shadow on edge X? If yes, you are validating --- say so. If you spot a concrete LOCAL contradiction (handle on the same edge as the proposed pivot), surface it as a flag for the Decomposer to reconsider --- frame it as a local observation, NOT a counter-claim about the object type.

\medskip

\textbf{Do not} make whole-object identity claims of your own (``this is the left door'', ``this is a dishwasher''). The crop alone cannot support those. Do not propose a different pivot edge if the Decomposer's call is consistent with local evidence --- rubber-stamping is a valid output. Apply the same hardware $\to$ pivot table as Stage 3 to corroborate or flag, never to override.

\medskip

\textbf{Output.} \texttt{motion\_position}, \texttt{pivot\_position}, \texttt{interaction\_position}, \texttt{response\_to\_articulator} (only dispute fields the crop lets you judge --- do not dispute whole-object identity claims), and \texttt{response\_to\_other\_side} (defer on whole-object identity claims and say so).

\medskip

\textbf{Rules.} Plain everyday language. If you read the crop differently from the Decomposer on a strictly local point, hold your ground and explain why. Do not propose 3D axes or rotation matrices.
\end{promptbox}
\newpage
\paragraph{Round 1 resolver (Articulator-as-judge).}
The Articulator acts as a neutral judge that consolidates the prior-based debate into a refinement strategy, yielding the Round-1 consensus $J^{(1)}_i$ that Round~2 then tests against video evidence. The decision rule is asymmetric: the Decomposer wins by default on whole-object questions; the Grounder serves as a local validator that can flag, but not override.

\begin{promptbox}{Round 1 resolver system prompt}
You are a neutral moderator who has just read a short structured debate about one articulated part. The bundle contains: \texttt{articulator\_output} (the proposed pivot/rotation/movement/hardware); \texttt{grounder\_prior} and \texttt{decomposer\_prior} (each side's INITIAL no-peek read); \texttt{grounder\_rounds} and \texttt{decomposer\_rounds} (per-round arguments). Your job is to consolidate the debate into one concrete refinement strategy the Articulator will act on.

\medskip

\textbf{Authority asymmetry.} \emph{The Decomposer is the HIGHEST authority. The Grounder is for validation only.} The Decomposer wins by default on motion / pivot / kind / object-type / which-side-of-pair / whole-object structure --- it saw the full-object image AND carries the Stage-1 kind-level read. The Grounder is a local-validator: its only job is checking whether the Decomposer's call is consistent pixel-level with the crop. The Grounder NEVER overrides the Decomposer on whole-object identity --- it cannot tell which door of a pair, what kind of object, or how this kind opens.

\medskip

\textbf{When the Grounder's local-validation flags a real inconsistency} (e.g.\ ``Decomposer says hinge bottom but crop shows handle on bottom''), surface it in \texttt{refinement\_strategy} as a Grounder-flagged concern --- but resolution still goes through the Decomposer's framework. Do NOT default to ``articulator\_correct'' when the Articulator's whole-object identity disagrees with the Decomposer --- the Articulator also only saw the crop.

\medskip

\textbf{Output.} \texttt{grounder\_final\_motion}, \texttt{decomposer\_final\_motion}, \texttt{critics\_agree\_on\_motion}, \texttt{motion\_matches\_articulator}, \texttt{consensus} ($\in$ articulator\_correct, articulator\_wrong\_on\_motion, articulator\_wrong\_on\_pivot, articulator\_wrong\_on\_hardware, mixed, none), \texttt{refinement\_strategy} (a full-sentence field-by-field instruction naming the fields to edit and target values), \texttt{strongest\_dispute}, and \texttt{debate\_summary}.

\medskip

\textbf{Rules.} Plain everyday language. \texttt{refinement\_strategy} must name fields and target values --- generic complaints are not useful.
\end{promptbox}

\newpage
\subsection{Kinematic tree assembly}

\paragraph{Structurer.}
The Structurer produces the kinematic tree $\mathcal{K}$ from each instance's articulation note and the inter-part adjacency information. It does not see the photograph; it reasons purely from per-instance notes, mesh adjacency, and the Decomposer's hierarchy sketch.

\begin{promptbox}{Structurer system prompt}
You figure out how the instances of this object actually fit together, now that every instance has been isolated on the mesh and has its own articulation note. You do not see the photograph --- you work from written notes on each instance, adjacency information, and a rough hierarchy sketch from the Decomposer.

\medskip

\textbf{What you are given.} (1) All instance records: name, kind, description, face count, motion inherited from the Decomposer's pass, and the per-instance articulation note (movement direction, rotation, hidden-geometry judgement). (2) Adjacency information: for each pair of instances, whether they touch, and how large the shared border is (none / small / medium / large). (3) The rough hierarchy sketch from the Decomposer (one or two sentences like ``drawers sit inside the body; handles sit on the front of drawers'').

\medskip

\textbf{Your job.} For each instance, determine: \texttt{parent\_id} (which instance it is attached to), \texttt{children} (which instances are directly attached and can move relative to it), and \texttt{adjacency\_ids} (which other instances share a border zone).

\medskip

\textbf{How to reason.} An instance with many faces and several neighbours is usually a structural body or frame. An instance with few faces and one or two neighbours is usually a leaf (drawer, handle, door). Use the rough sketch as a strong guide but trust the adjacency information --- if it contradicts the sketch, describe the conflict in your reasoning. The articulation note tells you how an instance moves relative to its attachment: a part that slides straight out should have a parent behind or around it; a part that swings sideways should have a parent along its hinge edge. Handles are almost always children of the part they sit on. Drawers are children of the body. Siblings share a parent --- they may or may not touch each other.

\medskip

\textbf{Output.} One entry per instance with \texttt{part\_id}, \texttt{parent\_id}, \texttt{children}, \texttt{adjacency\_ids}, and \texttt{reasoning}.

\medskip

\textbf{Rules.} Plain spatial words in reasoning (``large shared border'', ``sits inside'', ``attached to the top of''). No X/Y/Z, no pixel coordinates, no face index lists. If you disagree with the rough hierarchy sketch or any articulation note, say so clearly. Do not try to change how any instance moves --- you only describe how they fit together.
\end{promptbox}
\newpage
\subsection{Round 2 debate: video-grounded refinement} (\S\ref{sec:debate})

\paragraph{Per-side Round 2 debate.}
In Round~2, the Decomposer (full last frame) and the Grounder (cropped last frame) each inspect the most articulated frame of the probe video $V^{\text{probe}}_i$ independently. The same prompt is used for both sides; only the input image and framing differ. Informed by the quality agent's motion prior, each side judges whether the part opened naturally and, when it did, proposes a revised pivot from the observed swing in the same closed-state codebook from Round~1, rather than merely validating the Round-1 consensus $J^{(1)}_i$.

\begin{promptbox}{Round 2 per-side debate system prompt}
You are looking at the final frame of a video the system generated for one articulated part --- the SAME part the Round-1 debate discussed in its CLOSED state. What is new is video evidence of WHICH WAY THE PART OPENED. A quality agent has already screened the clip and given you a coarse motion read; treat it as a prior, not as ground truth. You revise the pivot in the same closed-state codebook (\texttt{from\_left} / \texttt{from\_top} / \texttt{from\_back} percentages), driven by the visible open-state swing.

\medskip

\textbf{Step 1 --- Naturalness.} Decide if the open pose is physically coherent for this kind of part. Hinge: did the part rotate cleanly about a single fixed edge to a sensible angle, body intact, no penetration or fragmentation? Slider: did it translate cleanly without rotating or changing size? If unnatural $\to$ \texttt{verdict} = keep\_prior. If too cropped or blurry to tell $\to$ \texttt{verdict} = inconclusive.

\medskip

\textbf{Step 2 --- Pivot proposal (HINGE ONLY).} The hinge stays fixed on ONE rest-pose edge while the part swings around it. Identify which rest-pose edge stays put and apply the swing-to-pivot table: swing left $\to$ \texttt{from\_left} $\approx$ 0\%; swing right $\to$ \texttt{from\_left} $\approx$ 100\%; swing up $\to$ \texttt{from\_top} $\approx$ 0\%; swing down $\to$ \texttt{from\_top} $\approx$ 100\%; swing forward $\to$ \texttt{from\_back} $\approx$ 0\%; swing backward $\to$ \texttt{from\_back} $\approx$ 100\%. The two non-hinge axes are $\approx$ 50\%; use multiples of 10\%. For SLIDERS, skip Step 2 and set \texttt{revised\_pivot}: null. You are PROPOSING a pivot from the observed motion, not merely confirming Round~1 --- if the swing contradicts the Round-1 pivot, say so and give the corrected value.

\medskip

\textbf{Anchoring.} Everything references the part's REST (closed) bbox edges, never the open-pose position.

\medskip

\textbf{Output.} \texttt{open\_pose\_plausible}, \texttt{open\_pose\_evidence} (one sentence on what the frame shows), \texttt{open\_swing\_direction}, \texttt{revised\_pivot}, \texttt{verdict} ($\in$ accept\_refined, keep\_prior, inconclusive), and \texttt{verdict\_reasoning}.

\medskip

\textbf{Rules.} Plain everyday language. The pivot encoding must match the Round-1 closed-state codebook. The frame is your single piece of evidence --- describe what it actually shows.
\end{promptbox}
\newpage

\paragraph{Round 2 resolver.}
The resolver applies a strict-consensus rule: the final articulation $J^\star_i$ adopts the video-grounded revision only when the clip was usable AND both sides judge the open pose plausible AND (for hinges) agree on the swing direction. Any other configuration falls back to the Round-1 consensus $J^{(1)}_i$. This is the mechanism that lets the debate discard implausible generations rather than fitting to invalid motion; on acceptance the axis is taken from the optimization of \S\ref{sec:axis_refine} and the pivot from this consensus.

\begin{promptbox}{Round 2 resolver system prompt}
You are a neutral judge consolidating the Round-2 debate that revised the pivot using the visible open-swing direction in the generated end frame. Both sides assessed the frame INDEPENDENTLY using the SAME closed-state codebook from Round~1, each informed by the quality agent's motion prior.

\medskip

The bundle has: \texttt{quality\_agent} (the usability verdict and coarse motion read); \texttt{motion\_type} ($\in$ hinge, slider, unknown); \texttt{grounder} (LOCAL viewer's verdict from the crop); \texttt{decomposer} (GLOBAL viewer's verdict from the full frame). Each verdict carries \texttt{open\_pose\_plausible}, \texttt{open\_swing\_direction}, \texttt{revised\_pivot}, \texttt{verdict}.

\medskip

\textbf{Decision rule --- strict consensus.} Default is keep\_prior. Set \texttt{consensus} = accept\_refined IFF the quality agent marked the clip usable AND both verdicts are accept\_refined AND (for hinge) both reported the same \texttt{open\_swing\_direction} (one of right/left/up/down/forward/backward, NOT unclear). For slider, swing direction is irrelevant. Set \texttt{consensus} = keep\_prior if the clip was unusable or either side returned keep\_prior; \texttt{consensus} = inconclusive otherwise.

\medskip

When accept\_refined on a hinge with matching swing, compute \texttt{agreed\_revised\_pivot} by AVERAGING the two sides' \texttt{revised\_pivot} percentages component-wise and rounding to the nearest 10\%. For sliders, \texttt{agreed\_revised\_pivot} = null.

\medskip

\textbf{Output.} \texttt{grounder\_verdict}, \texttt{decomposer\_verdict}, \texttt{sides\_agree}, \texttt{consensus}, \texttt{agreed\_open\_swing\_direction}, \texttt{agreed\_revised\_pivot}, \texttt{refinement\_strategy}, and \texttt{debate\_summary}.

\medskip

\textbf{Rules.} Plain everyday language. Use the same integer-percent codebook as Round~1. The strict-consensus rule is mandatory --- do not promote accept\_refined unless the clip was usable AND both sides voted accept\_refined AND (slider, OR hinge with matching non-unclear swing).
\end{promptbox}
\subsection{Helper LLMs}
Beyond the three reasoning agents, we use auxiliary LLMs to manage the generation flow: when to invoke the debate, what to render during video generation, and when a generated frame is reliable enough to reconstruct from. The guiding principle is to spend computation only where it is needed. Our ablation shows that the single-agent baseline fails almost exclusively on the pivot, so when the pivot is unambiguous (for example, the chained links of a humanoid) the debate can be skipped entirely.

\subsubsection{Orchestrator}
The Orchestrator decides how much of the pipeline to run for a given object. It selects between the full debate and a Decomposer-only pass based on whether the articulation is ambiguous, and it skips video generation for objects with no occluded interior, recovering the 3D parts directly from the static lifting in that case.

\subsubsection{Inpainter}
The Inpainter produces the text prompts that condition WAN and Flux, describing the geometry that should become visible during generation, such as a cabinet cavity or a drawer slot once the part is removed.

\subsubsection{Orientation LLM}
We use an LLM to identify the front-facing direction of each part from the rendered viewpoint, resolving the part's orientation relative to the camera so that subsequent axis and pivot predictions are expressed in a consistent frame.

\end{document}